\documentclass{article}

\usepackage{microtype}
\usepackage{graphicx}
\usepackage{subfigure}
\usepackage{booktabs} % for professional tables

\usepackage{hyperref}

\usepackage[accepted]{whi2018}

\graphicspath{{fig/}}

\icmltitlerunning{Interpretable Discovery in Large Image Data Sets}

\begin{document}

\twocolumn[
\icmltitle{Interpretable Discovery in Large Image Data Sets}



\icmlsetsymbol{equal}{*}

\begin{icmlauthorlist}
\icmlauthor{Kiri L. Wagstaff}{jpl}
\icmlauthor{Jake Lee}{col}
\end{icmlauthorlist}

\icmlaffiliation{jpl}{Jet Propulsion Laboratory, California Institute
  of Technology, Pasadena, CA, USA}
\icmlaffiliation{col}{Columbia University, New York, NY, USA}

\icmlcorrespondingauthor{Kiri L. Wagstaff}{kiri.l.wagstaff@jpl.nasa.gov}
%\icmlcorrespondingauthor{Eee Pppp}{ep@eden.co.uk}

\icmlkeywords{Machine Learning, ICML, Discovery, Convolutional Neural
  Networks, Interpretability}

\vskip 0.3in
]



\printAffiliationsAndNotice{\icmlEqualContribution} % otherwise use the standard text.

\begin{abstract}
%{\bf [revise at the end]}
Automated detection of new, interesting, unusual, or anomalous images
within large data sets has great value for applications from surveillance
(e.g., airport security) to science (observations that don't fit a
given theory can lead to new discoveries).  
%In particular, novelty
%detection in image data sets could help detect new near-Earth
%asteroids, fresh impact craters on Mars, and other key phenomena that
%might otherwise be lost within a large archive.  
Many image data analysis systems are turning to convolutional neural
networks (CNNs) to represent image content due to their success in
achieving high classification accuracy rates.  However, CNN
representations are notoriously difficult for humans to interpret.  We
describe a new strategy that combines novelty detection with CNN image
features to achieve rapid discovery with interpretable explanations of
novel image content.  We applied this technique to familiar images from
ImageNet as well as to a scientific image collection from planetary
science. % and ecology.  
%Finally, we conducted a user study to assess the
%utility of the generated explanations.
\end{abstract}

\section{Introduction}
\label{intro}

%Furthermore, please make sure that files contain only embedded Type-1 fonts
%(e.g.,~using the program \texttt{pdffonts} in linux or using
%File/DocumentProperties/Fonts in Acrobat). Other fonts (like Type-3)
%might come from graphics files imported into the document.

%The style file uses the \texttt{hyperref} package to make clickable
%links in documents. If this causes problems for you, add
%\texttt{nohyperref} as one of the options to the \texttt{icml2018}
%usepackage statement.

As more and more data is collected by science, industry, finance, and
other fields, the need increases for automated methods to identify
content of interest.  The discovery of new or unusual observations
within large data sets is a key element of the scientific process, since
unexpected observations can inspire revisions to current knowledge and
overturn existing theories~\cite{kuhn:science62}.  When exploring a
new environment, such as the deep ocean or the surface of Mars, quickly
identifying observations that do not fit our expectations is essential
for making the best use of limited mission lifetimes.  The challenge
is particularly acute for image data sets that may contain millions
of images (or more), rendering exhaustive manual review infeasible.

Many anomaly and novelty detection methods are available, but in
isolation their results can be difficult to interpret.  Once an
observation is identified as novel or anomalous, the next question is
generally, ``Why?''  To investigate the anomaly, users need to know
what properties of the observation caused it to be selected.  For
images, these properties might include color, shape, location, objects,
content, etc.

\begin{figure}
\begin{center}
\subfigure[Selection]{\includegraphics[width=1in]{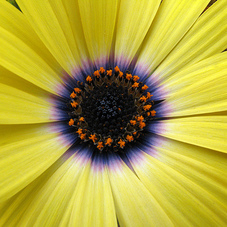}}
\subfigure[Expected]{\includegraphics[width=1in]{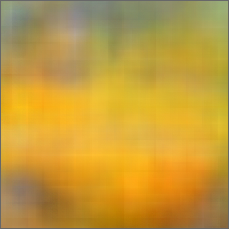}}
\subfigure[Novel]{\includegraphics[width=1in]{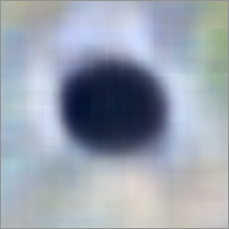}}
%\subfigure[Selection]{\includegraphics[width=1in]{fig/exp-yellow/02-select.jpg}}
%\subfigure[Expected]{\includegraphics[width=1in]{fig/exp-yellow/02-doso-recon.png}}
%\subfigure[Novel]{\includegraphics[width=1in]{fig/exp-yellow/02-doso-resid.png}}
%\subfigure[Selection]{\includegraphics[width=1in]{fig/exp-yellow/08-select.jpg}}
%\subfigure[Expected]{\includegraphics[width=1in]{fig/exp-yellow/08-doso-recon.png}}
%\subfigure[Novel]{\includegraphics[width=1in]{fig/exp-yellow/08-doso-resid.png}}
%\includegraphics[width=1.5in]{fig/study/02-heatmap-overlay.jpg}\\
\end{center}
\caption{Given a collection of yellow object images, this flower (a) was
  selected as novel.  The explanation consists of the ``expected
  content'' (b) and ``novel content'' (c). The yellow color
  is not novel, but the dark center is. }
\label{fig:exp}
\end{figure}

% contributions
In this work, we propose the first method to generate
human-comprehensible (visual) explanations for novel discoveries in
large image data sets.  A simple example is shown in
Figure~\ref{fig:exp}, where a flower with a dark center was selected
from a set of images containing yellow objects (e.g., banana, squash,
lemon).  The yellow color is not novel (middle), so it is omitted from
the explanation (right).  Instead, the dark center of the flower is
highlighted as novel.
We describe the details of how the selections are made and how the
explanations are generated, then conduct experiments that
include studies of well known ImageNet images and data compiled
for real scientific investigations.
%... applications to planetary science and biology...
%We also report on a user study that assessed whether the explanations
%successfully helped users understand why individual images were
%selected.
%
One of the urgent questions within the field is whether
interpretability has to come at the expense of accuracy or
performance.  Our results support an optimistic answer: in
the case of novel image detection, we can obtain explainable results
in tandem with the best discovery performance.

%{\bf figure: image, known content (reconstruction), novel content
%  (residual), heatmap]}

% LocalWords:  pdffonts linux hyperref clickable nohyperref icml usepackage XAI
% LocalWords:  todo Interp Explainability ImageNet heatmap

\section{Related Work}
\label{relwork}

%- (rare) class / novelty discovery

%- interpretable / explainable ML
Recently there has been a growing interest in interpretable or
explainable machine learning methods, especially for supervised
learning~\cite{biran:expl-survey17}.
Some methods are fundamentally interpretable, such as decision trees,
while others train a simplified ``mimic'' or approximation model that
provides a post-hoc rationalization for a given decision (e.g.,
LIME~\cite{ribeiro:lime16}).
%The LIME (Local Interpretable Model-Agnostic Explanations) method
%constructs a simplified model to produce explanations for decisions
%made by an existing, more complex model~\cite{ribeiro:lime16}.  LIME
%generates a locally linear approximation to a learned discriminative
%model (e.g., classifier) by selecting K features that will best
%separate the item under consideration from nearby items of different
%classes.  %Experiments showed that...
%Similarly, Hara and Hayashi~\yrcite{hara:interp16} build simplified
%``mimic'' models for random forests.
%
%- for images
Image classification explanations often take the form of a saliency
map that identifies the parts of an image that were relevant to the
classification decision.
Recent advances include a single-pass salience map generator that can
run in real-time~\cite{dabkowski:salience17}. 
Park et al.~\yrcite{park:attentive17} developed a Pointing and
Justification (PJ-X) model to answer questions about the content of an
image.  It provides a text explanation and an annotated image that 
highlights the image elements that led to the classifier's decision.

Fewer methods exist for generating explanations for unsupervised
learning methods.
Brinton proposed an Explainable Principal Components Analysis method
that uses human interaction to generate human-comprehensible principal
component vectors~\cite{brinton:epca17}.  This approach aims for an
explainable model, not explainable decisions.
Siddiqui et al.~\yrcite{siddiqui:seqfeatexp15} generated explanations
to help human experts determine whether a selected item is anomalous
or not.  Their approach incrementally reveals feature values until the
expert is sufficiently confident.  
%The explanations, therefore,
%consist of a ranked list of feature/value pairs.  
%The explanations
%were evaluated in terms of their parsimony (i.e.,~the number of
%features needed for the user to reach a confident conclusion).
%
The DEMUD algorithm uses a Singular Value Decomposition (SVD) model to
discover new classes and provide a custom explanation for each
discovery~\cite{wagstaff:demud13}.  
% DEMUD incrementally constructs a
% growing model of the user's knowledge and selects items with large
%reconstruction error (those most likely to represent new classes) for
%the user to review.  
The residual vector (information not captured by
the model) is provided as an explanation for the item's selection.
To our knowledge, no methods exist that generate visually meaningful
explanations for class discovery in image data sets.

\comment{
our work:
- we generate a justification (why selection was a good one) not an
explanation (how it was done)?
%- Barzilay et al 1998 - rule-based systems - 3 layers - domain,
%strategic, and communication
%- Lazaridou et al. 2017? - neural agents learn to comm with each other
%about images
- human-grounded evaluation (ability to reason about a model) -
Doshi-Velez and Kim, 2017
  - application-grounded, human-grounded, functionally-grounded (proxy)
%Finale Doshi-Velez and Been Kim. A roadmap for a rigorous science of
%interpretability. arXiv preprint arXiv:1702.08608, 2017.
- vis hidden states in CNNs - Simonyan et al., 2013; Zeiler and Fer-
gus, 2013
%Karen Simonyan, Andrea Vedaldi, and Andrew Zisserman. Deep inside
%convolutional networks: Visualising image classification models and
%saliency maps. CoRR, abs/1312.6034, 2013.
%Matthew D. Zeiler and Rob Fer- gus. Visualizing and understanding
%convolutional net- works. CoRR, abs/1311.2901, 2013.
- biran and mckeown 2017 - show intersection of feature's expected and
actual contrib - can show missing info as well as extra; NLG of
explanations/justifications 
%Or Biran and Kathleen McKe- own. Human-centric justification of
%machine learning pre- dictions. In IJCAI, Melbourne, Australia, 2017.
}

% LocalWords:  Hara Hayashi hara interp al PJ Brinton Siddiqui siddiqui DEMUD
% LocalWords:  seqfeatexp SVD Barzilay Lazaridou comm Doshi Velez roadmap arXiv
% LocalWords:  Simonyan Zeiler Fer gus Andrea Vedaldi Zisserman Visualising abs
% LocalWords:  CoRR biran mckeown feature's contrib NLG McKe centric pre IJCAI

% approach
\section{Visual Explanations for Novelty Detection}

\begin{figure}
\begin{center}
\includegraphics[width=3.2in]{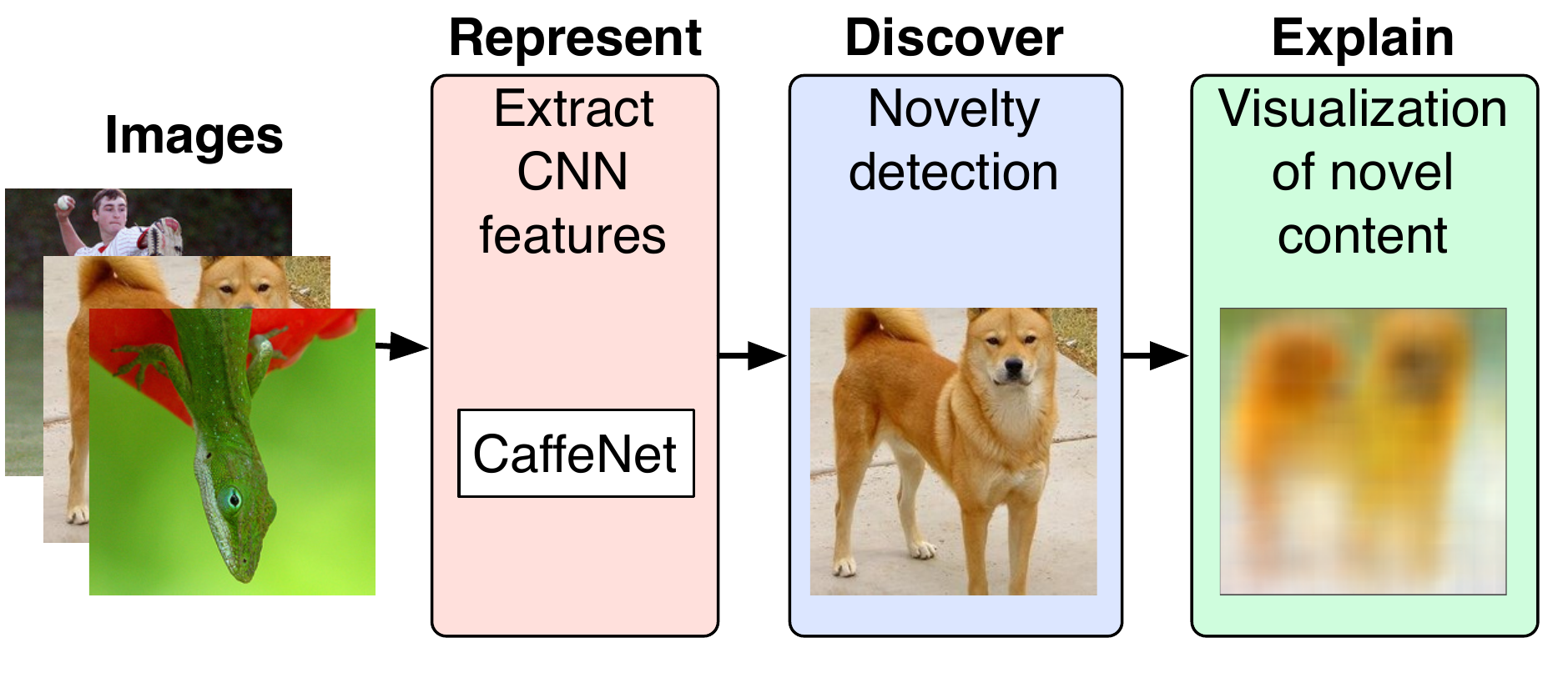}
\end{center}
\caption{Interpetable image discovery system}
\label{fig:sys}
\end{figure}

Our approach to interpretable discovery in image data sets uses a
novelty detection algorithm to select images and generate raw
explanations, a convolutional neural network (CNN) to represent
abstract image content, and CNN feature visualization methods to
render the explanations understandable to humans (see
Figure~\ref{fig:sys}).  
%Our implementation is available at 
%\url{http://withheld.for.blind.review/}.

\subsection{Novelty Detection with Explanations}
%- Use DEMUD to iteratively select the most unusual images

To detect novel images within a data set, we employed the DEMUD
algorithm~\cite{wagstaff:demud13}, which automatically generates
explanations during the novelty detection process.
DEMUD incrementally builds an SVD model of what is known about a data
set $\mathbf{X}$.  It proceeds by iteratively selecting the most
interesting remaining item, with respect to the SVD model, and then
incorporating it into the SVD model.  Interestingness (or novelty) is
estimated using reconstruction error, where a higher error indicates more
novelty.  Reconstruction error $R$ for each item $\mathbf{x}$ is
calculated as
\begin{equation}
R(\mathbf{x}) = ||\mathbf{x} - (\mathbf{U}\mathbf{U}^T (\mathbf{x} - \mu) +
\mu) ||_2,
\end{equation}
where $\mathbf{U}$ is the current set of top $K$ eigenvectors from the
SVD of $\mathbf{X_s}$, the set of already selected items, and $\mu$ is the mean
of all previously seen $\mathbf{x} \in \mathbf{X_s}$.  The most
interesting item $\mathbf{x}' = \mbox{argmax}_{\mathbf{x} \in \mathbf{X}} \:
R(\mathbf{x})$ is moved from $\mathbf{X}$ to $\mathbf{X_s}$, and an
incremental SVD algorithm updates $\mathbf{U}$ to incorporate $\mathbf{x}'$.
This approach minimizes redundancy in the selections, since items
similar to those previously selected will have low reconstruction
error.  
%DEMUD was shown to out-perform existing supervised and
%unsupervised methods for class discovery, including a standard SVD
%approach that models the whole data set and looks for
%outliers~\cite{wagstaff:demud13}.

DEMUD's explanation for each selection $\mathbf{x}'$ is the residual
difference between a reconstructed $\mathbf{x}'$, after projection
into the low-dimensional space defined by $\mathbf{U}$, and its
original values.  The reconstruction of $\mathbf{x}'$ is
\begin{equation}
\hat\mathbf{x}' =  \mathbf{U}\mathbf{U}^T(\mathbf{x}' - \mu) + \mu 
\end{equation}
and the explanation $e$ is $e = \mathbf{x}' - \hat\mathbf{x}'$. 
The explanation captures the information contained in $\mathbf{x'}$
that the current model $\mathbf{U}$ could not represent.

%A side benefit of DEMUD is that the
%SVD residuals provide an explanation for each item that can be
%examined to understand what features/values led to its selection.

To our knowledge, our work is the first attempt to apply DEMUD to
image data and obtain meaningful explanations.  The straightforward
approach of providing DEMUD with the pixel values in the image does
not perform well with respect to image categories
because it is overly sensitive to small changes in position,
illumination, etc.  Further, the explanations that are generated by
DEMUD are a sequence of pixel values that are very difficult to
interpret.  An interest in the image content, rather than its pixel
values, calls for a more abstract representation.  We solve this
problem by employing a convolutional neural network to represent each
image prior to novelty detection.

\subsection{CNN Features for Image Content Representation}
%- Extract CNN features from images in data set

Several methods exist for extracting image content, such as
LBP~\cite{ojala:lbp02}, SIFT~\cite{lowe:sift04}, and
HOG~\cite{dalal:hog05}. Recently, the representation of image content
provided by trained convolutional networks has been shown to be useful
for a variety of visual tasks, not just the original classification
task for which the network was trained~\cite{razavian:CNNfeat14}.  
% todo:
%\cite{zhou:CNNlocalize15}
%\cite{yosinski:deepvis}

We extract a feature vector to represent each image by propagating the
image through a trained neural network.  Any network (or autoencoder)
trained on a sufficiently diverse set of inputs could be employed.
For our experiments, we used CaffeNet~\cite{jia:caffe14}, a version 
of AlexNet~\cite{krizhevsky:alexnet12} that was trained on 1.2 million
images from 1000 classes in the ImageNet data set.  For each image, we
%resized our input images to 227 x 227 pixels, 
%propagated them through the CNN, and
recorded the activations at each fully connected layer.

\subsection{Visualization of Explanations}
%- Convert DEMUD residuals to human-understandable explanations [key contrib]

\comment{
DEMUD builds an SVD of the data matrix of previously selected items
$\mathbf{X_S}$ by estimating the principal component vectors
$\mathbf{U}$ from $\mathbf{X_S}^T = \mathbf{U} \mathbf{\Sigma}
\mathbf{V}^T$.  For each remaining item $\mathbf{x}$ in the full set
$\mathbf{X}$, DEMUD calculates the reconstruction error between
$\mathbf{x}$ and $\hat{\mathbf{x}}$, the reconstruction of $\mathbf{x}$:
\begin{eqnarray}
R(\mathbf{x}) &=& ||\mathbf{x} - \hat{\mathbf{x}}||_2 \\
\label{eqn:r}  
&=& ||\mathbf{x} - (\mathbf{U}\mathbf{U}^T (\mathbf{x} - \mu) +
\mu) ||_2, %\\
\end{eqnarray}
where $\mu$ is the mean of all previously seen $\mathbf{x} \in
\mathbf{X_S}$.  The item $\mathbf{x'}$ with the largest reconstruction
error is selected.  The ``explanation'' for selecting $\mathbf{x'}$ is
the residual vector $\mathbf{x'} - \hat{\mathbf{x}}'$, i.e., the
information contained in $\mathbf{x'}$ that the current model
$\mathbf{U}$ could not represent.
}

When previous researchers applied DEMUD to numeric data sets, the
residuals could be directly interpreted because each feature already
represented a human-comprehensible value (e.g., size, age, number of
petals).  In contrast, in the image domain, residual values for the
4096 features employed by layer fc6 in CaffeNet are not directly
interpretable.  Note that visualizing these externally generated
feature vectors is not addressed by methods that seek to visualize the
learned CNN model itself, such as that of Zeiler and
Fergus~\yrcite{zeiler:viscnn14}, or to generate synthetic inputs to
visualize class membership such as DeepVis~\cite{yosinski:deepvis15}.  

Two recent advances provide methods for visualizing CNN feature
vectors as images.  The Deep Goggle (DG) system employs gradient
descent to generate a synthetic input image that yields the same
layer-level activation as a given
target~\cite{mahendran:deepgoggle15}. Dosovitskiy and
Brox~\yrcite{dosovitskiy:upconv16} trained an up-convolutional (UC)
network to take the layer-level activation as an input and predict the
corresponding original image.  DG tends to highlight fine
details, while UC more faithfully represents color and
location.  We adopted the UC method for this study.

\begin{figure}
\begin{center}
\subfigure[Selection]{\includegraphics[width=1in]{exp-yellow0-02-select.jpg}}
%\subfigure[Expected]{\includegraphics[width=1in]{fig/exp-yellow0/02-doso-recon.png}}
\subfigure[Pixel]{\includegraphics[width=1in]{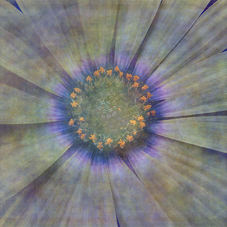}}
\subfigure[CNN]{\includegraphics[width=1in]{exp-yellow0-02-doso-resid.png}}
\end{center}
\caption{Yellow flower (a) with novelty explanations based on
  pixel-level features (b) versus CNN features (c).}
\label{fig:exp2}
\end{figure}

Previous investigators used these visualization methods
to generate images that correspond to the internal CNN representations
of other (real) images.  We instead employ UC to visualize the DEMUD
reconstruction $\hat{\mathbf{x}}'$ and residual $e$, which are not
real images.  These visualizations divide the image content
into what is expected ($\hat{\mathbf{x}}'$) and what is new ($e$), as
shown in Figure~\ref{fig:exp}.
%Our hypothesis is that the visualization of the residual
%information can serve to highlight what visual content is novel about
%the selected image.
%{\bf [Explain heatmap generation]}
%{\bf [Show an example]}
%An example was shown in Figure~\ref{fig:exp}.
Figure~\ref{fig:exp2} shows the same yellow flower image with a
visualization of the residual obtained when using pixel-level
features, in which almost every pixel value is highlighted as novel
and it is difficult to understand why the image was selected.
Figure~\ref{fig:exp2}(c) shows the explanation obtained using CNN
features from the first fully-connected layer of CaffeNet (fc6).  As
expected, fine details are not shown; the fc6 representation instead
captures higher level image content, and the image shown here includes
only the novel components (e.g.,~the dark center of the flower). 

% LocalWords:  Interpetable CNN DEMUD SVD argmax LBP todo autoencoder CaffeNet
% LocalWords:  ImageNet fc softmax contrib Zeiler Fergus zeiler viscnn DeepVis
% LocalWords:  DG Dosovitskiy Brox dosovitskiy upconv UC heatmap AlexNet

\section{Experimental Results}

We conducted several experiments to assess (1) the ability to discover
new classes in progressively more difficult conditions and (2) the
quality of the generated explanations. We present two here for the
sake of brevity. 
All data sets, extracted features, and evaluation scripts are
available at \url{http://jakehlee.github.io/interp-img-disc.html}.

%\subsection{Data Sets}

\begin{figure*}
\begin{center}
% Use width=2.22in if you want to fit 3 across
\subfigure[Balanced]{\includegraphics[width=3in]{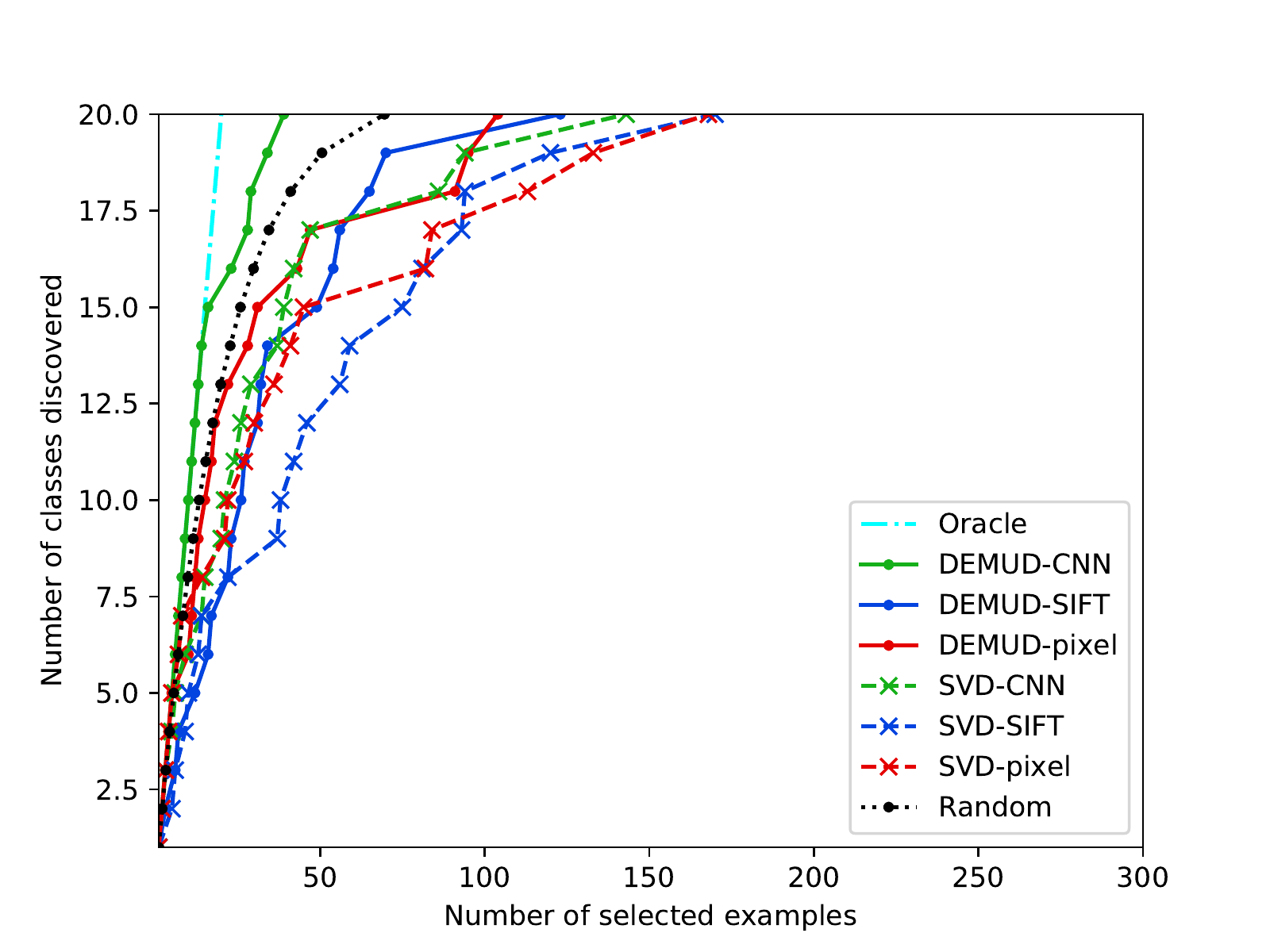}}
\subfigure[Unbalanced]{\includegraphics[width=3in]{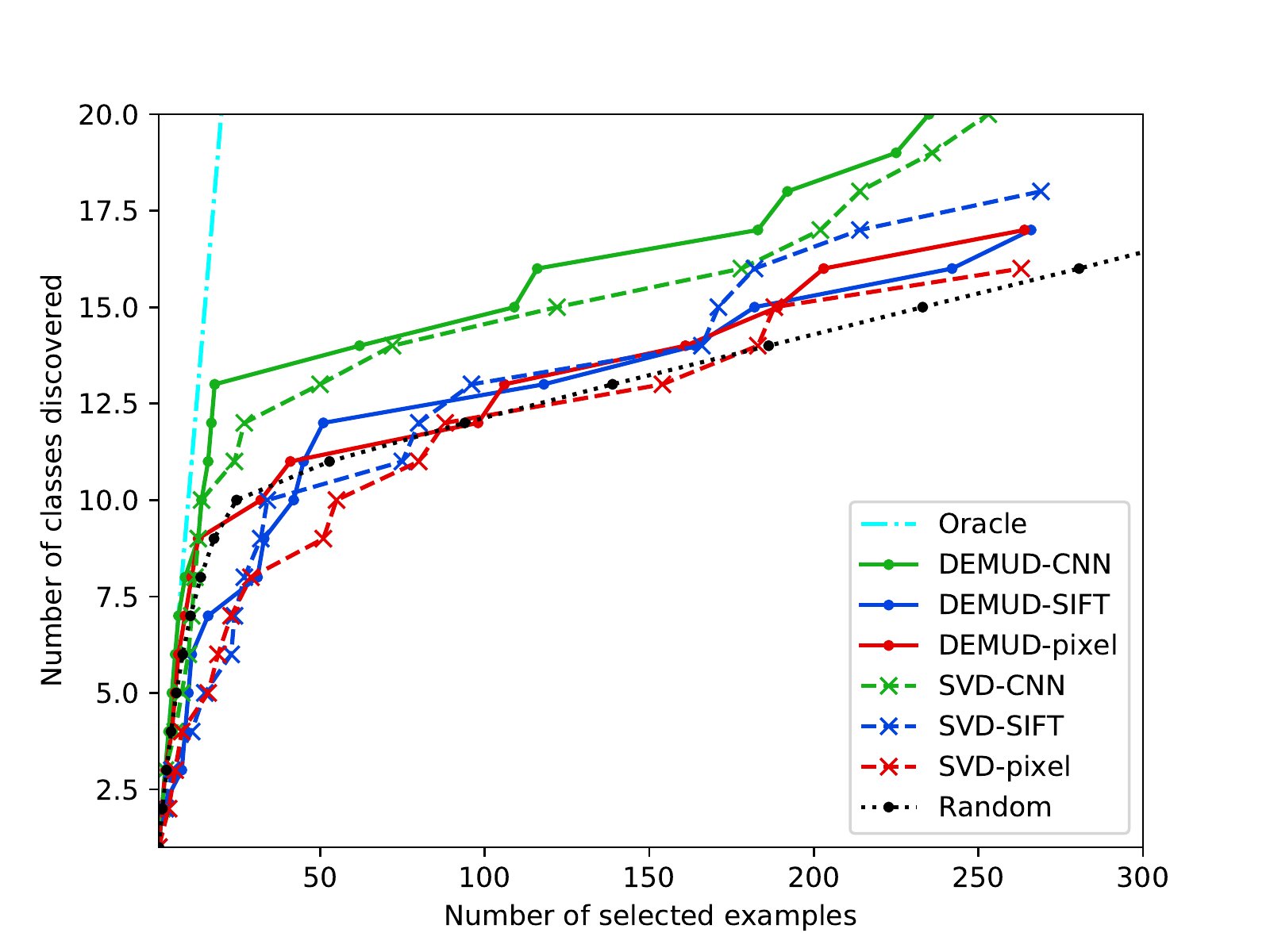}}
\end{center}
\caption{Discovery of 20 randomly chosen ImageNet
  classes.  DEMUD with CNN layer fc8 features achieved the best performance.}
\label{fig:in-perf}
\end{figure*}

\subsection{Methodology}
% and evaluation

We compared two discovery methods (DEMUD and a standard batch-mode SVD) using
three different representations: pixels, SIFT features, and CNN-based
features. 
%- Representations: pixels, CNN feat, SIFT keypoints?
%

%DEMUD requires the specification of an initial ``seed'' item.  The
%DEMUD software employs a heuristic that first performs a full SVD on
%the entire data set and selects the item with the largest
%reconstruction error.
%For each DEMUD experiment, we conducted 20 trials, each using a different
%randomly selected seed item.
%

To assess discovery, we used each algorithm to generate a ranked list
of images in descending order of novelty.  We used the image class
labels (which were not visible to the algorithms) to compute $C_i$,
the cumulative number of distinct classes  (e.g., dog, baseball, seal,
...) that were discovered up to selection $i$.  Plotting the number of
discovered classes as a function of the number of items selected
yields a discovery curve.  We calculated the normalized area under the
curve (nAUC), given $k$ classes, by summing $C_i$ from selection 1 to
$t$ and dividing by the AUC achieved by perfect discovery:
%.  Given $k$ classes in the data set, perfect discovery is achieved
%when a new class is discovered with each of the first $k$ selections:
\begin{equation}
nAUC_t = \frac{\sum_{i=1}^t C_i}{\sum_{i=1}^k i + (t-k)\times k} \times 100.0.
\end{equation}

In all DEMUD and SVD runs, we used a $K$ value (number of principal
components) of 50.  Results were not very sensitive to this choice;
sweeping $K$ from 10 to 100 yielded results with a standard deviation
of less than 1\% nAUC.
% see plot_rand_kvals.py: (k=10,20...100)
% fc6: mean 95.98, std 0.32
% fc7: mean 96.93, std 0.46
% fc8: mean 98.24, std 0.67
Random selection results are the average of 1000 trials.  All other
methods are deterministic.

We generated SIFT-based features using a visual bag of words approach.
We clustered SIFT keypoints across all images within a data set, then
represented each image with the distribution of its keypoints across
clusters.  Since there is no standard way to select the best number of
clusters in advance, for all SIFT results we report the best
performance achieved after testing ${3,4,5,10,15,20}$.
Note that this representation cannot provide meaningful explanations.
DEMUD residuals will be in the form of distributions of unexpected
values for keypoint cluster histograms, which have no
clear path to visualization.  However, we include SIFT as a comparison
for the discovery step given its ubiquity in image analysis work.

%\subsection{ImageNet Images}
\subsection{Discovery of ImageNet Classes}

%Our first experiment was conducted on a data set composed of randomly
%selected classes from the 1000 classes represented in the ImageNet
We first experimented with the ImageNet data set that was compiled for
the Large Scale Visual Recognition 
Challenge (ILSVRC) in 2012~\cite{imagenet12}.
% classes were:
% 133: American egret
% 168: English foxhound
% 629: liner, ocean liner
% 41: American chameleon
% 888: vestment (a chasuble or other robe worn by the clergy or
% choristers during services; did not know that)
% 261: chow, chow chow (Northern chinese dog, adorable. Very happy
% with this randomness.)
% 151: Sea lion
% 249: Eskimo dog, husky
% 742: prayer rug, prayer mat
% 789: shoe shop
% 111: flatworm
% 174: Ibizan hound
% 430: baseball
% 283: tiger cat
% 745: projectile, missile
% 596: harvester, reaper
% 909: wing
% 607: iron, smoothing iron
% 813: space shuttle  
% 758: recreational vehicle
We randomly selected 50 images from the ILSVRC12 training set for each
of 20 classes to obtain a total of 1000 images.  Each image was cropped as
needed to obtain a 1:1 aspect ratio, then resized to 227x227 pixels.
Figure~\ref{fig:in-perf}(a) shows the number of classes discovered as
a function of the number of selections.  The ``Oracle'' line shows
perfect discovery performance (i.e.,~a new class discovered with each
selection).  

%Random selection of images performed moderately well on this data set
%of 20 evenly distributed classes.  
DEMUD using the pixel values or SIFT features to represent the images
performed slightly worse than random selection in class discovery.  In
all cases, DEMUD out-performed the use of a standard SVD. The use of
CNN-based representations accelerated class discovery for both
methods. DEMUD using CNN layer fc8, which has the highest level
of content abstraction, had the highest performance.
In fact, it achieved perfect performance for the first 14 selections.

\paragraph{Unbalanced classes.}
For real discovery problems, classes are unlikely to be equally
balanced in the data set.
%; that is generally achievable only when the
%classes are known in advance.  
%Imbalanced classes present a more difficult
%challenge for discovery.  
We expected that random selection would
perform worse on an unbalanced data set, while DEMUD would be more robust
to class imbalance.  To test this hypothesis, we generated a variant
of the ImageNet data set in which the first 10 classes contain 50
items and the last 10 classes have only one item.  As shown in
Figure~\ref{fig:in-perf}(b), this data set was much more difficult:
the minority classes took much longer to discover.
However, there was a clear improvement when using CNN representations
versus using pixel or SIFT representations, and DEMUD again
out-performed a standard SVD. 

Table~\ref{tab:in-auc} compares results on the balanced and unbalanced
data sets, for DEMUD and the SVD, for each representation, and the
random selection baseline.  As expected, DEMUD suffered less of a
performance drop than was observed for random selection, and the CNN
features provided much more robustness than did SIFT or pixel features.

% table showing AUC for:
% DEMUD-CNN-fc[678], SVD-CNN-fc[678], DEMUD-pixel, SVD-pixel, random
% (9 values)
% cols: DEMUD, SVD | DEMUD-ubal, SVD-ubal
% rows: fc6, fc7, fc8, pixel, SIFT?
% random perf in caption; bold values better than random?
\begin{table}
\caption{Discovery nAUC$_{300}$ on 20 random ImageNet classes. %(20 trials,
%  (300 selections).%, $K=50$).  
The best result for each data set is in {\bf bold}.}
\label{tab:in-auc}
\begin{center}
\begin{tabular}{|l|cc|cc|} \hline
 & \multicolumn{2}{c}{Balanced} & \multicolumn{2}{|c|}{Unbalanced} \\
Features & DEMUD & SVD  & DEMUD & SVD \\ \hline
CNN-fc8 & {\bf 98.55} & 91.86 & {\bf 81.98} & 78.18 \\
CNN-fc7 & 97.11 & 92.44       & 71.50 & 69.26 \\
CNN-fc6 & 96.27 & 93.72       & 69.43 & 68.14 \\ \hline
SIFT    & 91.67 & 86.21 & 66.59 & 68.36 \\ \hline
Pixel   & 93.43 & 88.71       & 67.97 & 62.51 \\ \hline
Random sel. & \multicolumn{2}{c|}{96.35} & \multicolumn{2}{c|}{63.94} \\ \hline
\end{tabular}
\end{center}
\end{table}

\begin{figure}[h]
\begin{center}
\setlength\tabcolsep{2pt} % default value: 6pt
\begin{tabular}{ccc}
%Starting image & & \\
%\includegraphics[width=1in]{fig/exp-rand-ubal/00-seed.jpg} & & \\ 
Selection & Expected & Novel \\
% note: these are all from fc6k50
\includegraphics[width=1in]{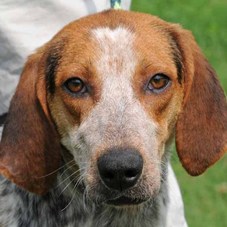} &
\includegraphics[width=1in]{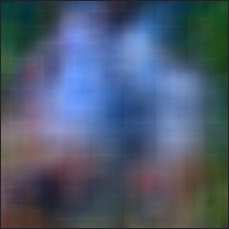} &
\includegraphics[width=1in]{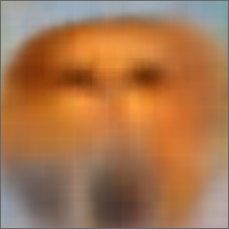} \\
\includegraphics[width=1in]{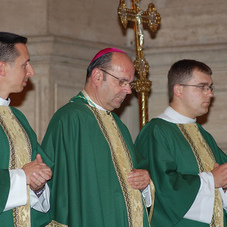} & 
\includegraphics[width=1in]{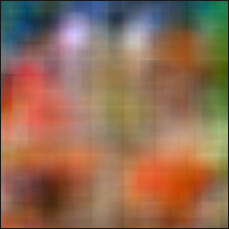} &
\includegraphics[width=1in]{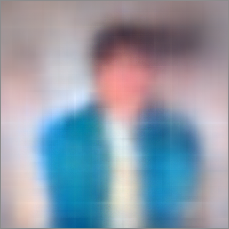} \\
\end{tabular}
\end{center}
\caption{Example DEMUD selections and explanations from random ImageNet
  classes.  Top: discovery of class ``English foxhound''.
%; green  background and white elements are not novel, but dog face shape and
%  orange color are.  
  Bottom: discovery of class ``vestment''.}% (and first standing human).}
%; note single person and blue tone of robe in ``novel'' image.} 
\label{fig:exp-rand}
\end{figure}

Figure~\ref{fig:exp-rand} shows two examples of novel selections and
their explanations. All visualizations are shown using the UC
method~\cite{dosovitskiy:upconv16}.
% for which only the fc6 network is
%available for the current version of Caffe.
% DEMUD selected the first image by performing an
%SVD across the entire data set and choosing the item with largest
%reconstruction error (from class ``chow chow'').  
DEMUD's first selection (not shown) was from the ``chow chow'' class and
includes a kneeling man with two small dogs on grass.  The top row
shows selection 2, which is the first discovery of an image from the
``English foxhound'' class.  The green grass background and white
elements are expected, but the shape of the dog's face and its orange
elements are novel.  The bottom row shows selection 16 (discovery of
class ``vestment''), at which point many of the colors and content are
expected.  However, it is the first image to include a standing human,
and the blue tones of the robe (which is bluer than grass) are
emphasized.  Note that the number and pose of the humans is abstracted
away in the explanation of novel content.

\subsection{Discovery in Mars Rover Images}

We applied the same techniques to a publicly available scientific
image data set. % to assess discovery with relevance to real
%investigations.  
This data set\footnote{\url{http://doi.org/10.5281/zenodo.1049137}}
consists of 6712 images from 25 classes that were collected by the
Curiosity Mars rover from sol (Martian day) 3 to 1060.
% ARGH!  This includes 'sun' class probably.  Should re-do with only
% the 6691 images actually posted at that URL.
The classes consist of the ground, horizon, and various rover parts
(e.g., wheel, drill, scoop).  The number of images per class ranges
from 21 to 2684.
%- closely related?

While the experiments in preceding sections used the same
ImageNet images that were used to train the CNN, the images in this
data set come from a very different distribution and have different
properties.  
If we use CaffeNet to predict the classes of these images, the most
common predictions are ``horned viper'', ``sandbar'', ``tick'',
``nematode'', and ``cliff dwelling'' (none of which are correct, of
course). 
This experiment therefore also tests whether the Earth-specific abstract 
concepts learned by CaffeNet generalize sufficiently well to a new
domain. 

%{\bf [Show figure with first 3 or so images + explanations]}

\begin{table}
\caption{Discovery nAUC$_{300}$ on Mars rover image data.}
%  $K=50$). 
%The best result for each data set is in {\bf bold}.}
\label{tab:sci-auc}
\begin{center}
\begin{tabular}{|l|cc|} \hline % |cc|} \hline
% & \multicolumn{2}{c}{Mars rover} \\ % & \multicolumn{2}{|c|}{Insects} \\
Features & DEMUD & SVD \\ \hline % & DEMUD & SVD \\ \hline
CNN-fc8 & 91.14 & 70.28 \\ % & {\bf 98.76} & 74.21 \\
CNN-fc7 & {\bf 92.75} & 74.60 \\ % & 96.81 & 81.27 \\
CNN-fc6 & 89.56 & 76.03 \\ \hline % & 86.00 & 68.32 \\ \hline
SIFT    & 68.47 & 18.56 \\ \hline % & 97.60 & 82.32 \\ \hline 
% MSL: SIFT k=15, static k=3
% stonefly9: SIFT k=5, static k=10
Pixel   & 69.17 & 55.36 \\ \hline % & 58.78 & 42.72 \\ \hline
Random sel. & \multicolumn{2}{c|}{76.33} \\ \hline % & \multicolumn{2}{c|}{97.45} \\ \hline
\end{tabular}
\end{center}
\end{table}

\begin{figure}
\begin{center}
%\subfigure{\includegraphics[width=3in]{fig/perfplot-msl.pdf}}
%\subfigure{\includegraphics[width=3in]{fig/perfplot-msl0-k50-fc6.pdf}}
\subfigure{\includegraphics[width=3in]{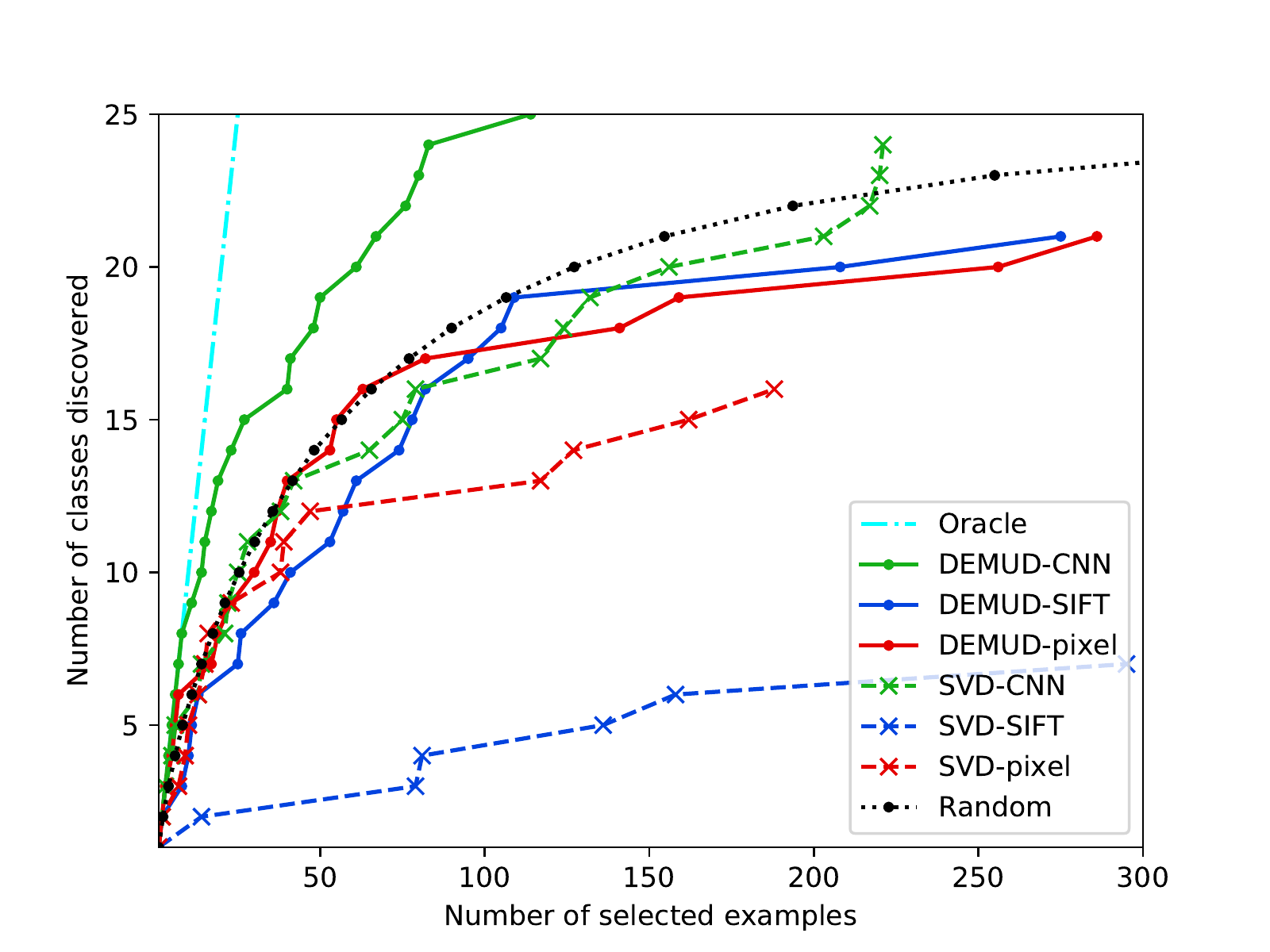}}
%\subfigure{\includegraphics[width=3in]{fig/perfplot-msl0-k50-fc8.pdf}}
\end{center}
\caption{Discovery of Mars rover image classes %(K = 50,
  (CNN layer fc7).}
\label{fig:msl-perf}
\end{figure}

As expected, this data set yielded lower discovery performance than
the ImageNet data sets (see Figure~\ref{fig:msl-perf} and
Table~\ref{tab:sci-auc}).  DEMUD-CNN was the only method to
perform better than random selection, which it exceeded by a large
margin.  After 300 selections, DEMUD-CNN was also the only method to
have discovered all 25 classes. Interestingly, on this data set, the
fc7 CNN layer provided the best representation (although the results
with fc8 were not very different).
%However, enough relevant content was captured by the CNN representations that
%DEMUD-CNN strongly out-performed the SIFT and pixel representations.

\begin{figure}[t]
\begin{center}
\setlength\tabcolsep{2pt} % default value: 6pt
\begin{tabular}{ccc}
Selection & Expected & Novel \\
\includegraphics[width=1in]{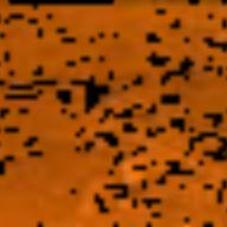} &
\includegraphics[width=1in]{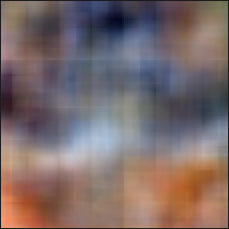} &
\includegraphics[width=1in]{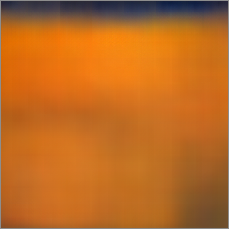} \\
\includegraphics[width=1in]{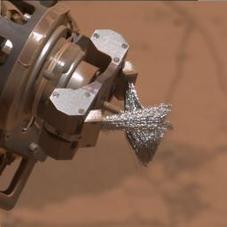} &
\includegraphics[width=1in]{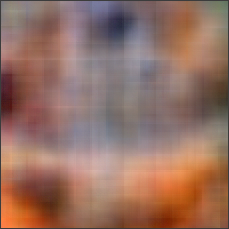} &
\includegraphics[width=1in]{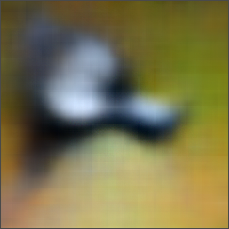} \\
\end{tabular}
\end{center}
\caption{Mars rover images: discovery of ``ground'' (top) and ``DRT side view''
  (bottom).} 
\label{fig:exp-msl}
\end{figure}

Figure~\ref{fig:exp-msl} shows some example selections %and
%explanations 
of interest.  The top row is the first discovery of the
``ground'' class (selection 5), where the orange component of the image is the
dominant part of the novel content.  All selections prior to this one
were close-ups of rover parts that did not include the bright orange
ground.  (This image also appears to be unusually saturated.)  The
bottom row shows the discovery of the ``DRT side view'' class
(selection 19; DRT is
the Dust Removal Tool or brush).  While the expected content is very
generic, the novel content visualization crisply highlights the tool
and brush tips.  Note that the orange component of the ground is no
longer novel; in fact, in comparison with the previously selected
``ground'' image, this terrain %appears washed out (with 
contains relatively less red, which manifests as more yellow/green in
the explanation. 

\begin{figure}[t]
\begin{center}
\setlength\tabcolsep{2pt} % default value: 6pt
\begin{tabular}{ccc}
%Sol 36 & Sol 808 & Novel (Sol 808) \\
Selection & Expected & Novel \\
\includegraphics[width=1in]{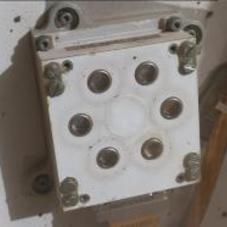} &
\includegraphics[width=1in]{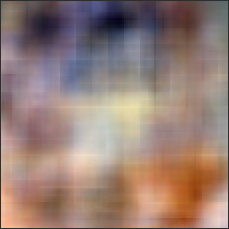} &
\includegraphics[width=1in]{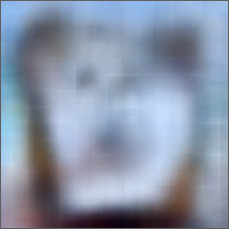} \\
\includegraphics[width=1in]{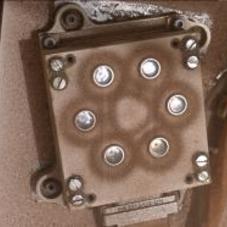} &
\includegraphics[width=1in]{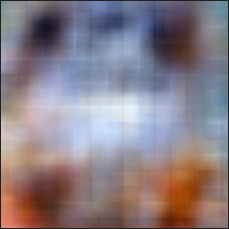} &
\includegraphics[width=1in]{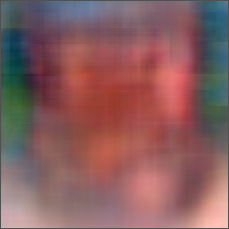} \\
\end{tabular}
\end{center}
\caption{Mars rover images: REMS-UV sensor first discovery (top)
  and later view when covered in dust (bottom, 772 sols later).}
\label{fig:exp-msl-remsuv}
\end{figure}

To see how the explanations aid interpretation, consider the images
shown in Figure~\ref{fig:exp-msl-remsuv}.  DEMUD discovered the
REMS-UV sensor in an image from sol 36 on selection 2 (due to its
shape), then selected an image of the same object more than 700 sols
later on selection 13 (due to its color). 
In terms of image class label, the second image is not a novel
discovery.  However, the explanation shows that selection 13 was
chosen because it is distinctly redder (more dusty) than expected,
which highlights the process of dust deposition on Mars over the
intervening time.  DEMUD did not have access to information about the
time ordering of the images; it naturally selected two extreme
examples (pristine and dust-covered) that emerged as novel within the
data set.

% How to use this tactically
Analyzing the full image archive is of interest for retrospective
studies, but this same approach can be useful in an immediate,
tactical setting.  Each day, a team of experts examines the latest
downlinked rover images to decide which targets the rover should examine
the next day.  DEMUD could be used to analyze the latest
batch of images and identify those that are novel with respect to the
archive.  By highlighting these images, DEMUD could help reduce the
chance that, in the high-pressure planning environment, a discovery
that merits follow-up study goes unnoticed. 

% LocalWords:  ImageNet DEMUD CNN fc SVD keypoints nAUC AUC rand kvals py std
% LocalWords:  OpenCV's keypoint ILSVRC chameleon vestment chasuble chinese sel
% LocalWords:  Ibizan cols ubal perf UC Caffe ccc AlexNet CaffeNet RGB ness DRT
% LocalWords:  tibetan Yorkshire ARGH sandbar MSL stonefly CaffeNet's warplane
% LocalWords:  Zapada Calineuria californica Hesperoperla pacifica Doshi Velez
% LocalWords:  doshi velez interpeval repr Heatmap Avg conf pre cc sol REMS UV
% LocalWords:  sols

\section{Conclusions and Future Work}

We have developed the first approach to generating visually meaningful
explanations for discovery in image data sets.  We employ a
convolutional neural network to generate abstract representations of
image content, then use the DEMUD algorithm to select novel images and
generate explanations in the form of residuals (information in the
image that could not be represented by the current model).  An
up-convolutional network generates human-interpretable visualizations
of the explanations.

In our experiments, we found that this approach achieved strong
(sometimes near-perfect) discovery performance, even in challenging
data sets.  DEMUD with CNN features always achieved higher discovery
performance than using pixel-based representations.  Further, 
% in all data sets but one, 
% ImageNet-Dogs - we would like to better understand this!
the use of CNN features consistently out-performed using SIFT.
Performance gains %(versus pixel representations) 
were most dramatic
for the Mars rover data set
% science investigation data sets from a Mars rover and an
% ecology investigation, 
in which the classes are imbalanced, the images are highly similar in
color, and pixel and SIFT representations are inadequate.
%
%Our user study found that the explanations increased user confidence
%in their understanding of why images were identified as novel.  
%We would like to continue refining this visualization to further
%improve confidence and trust in the system.

There are some potential limitations in the ability of a neural
network that was trained on ImageNet images to generalize to data sets
with very different properties, such as those from Mars.
% or insect studies.  
In future work, we will explore whether a new
network (or autoencoder) that was trained on images from the target
distribution would yield even better discovery performance.
We have also found that DEMUD is very good at detecting mislabeled
examples, since they contain unusual content with respect to their
class.  In an initial experiment with ImageNet classes, DEMUD
discovered a lion image (with mane) that was labeled as a ``tiger cub.''
This capability could be useful in exploring even fully labeled data
sets, to help identify labeling errors and/or adversarial examples.

This approach can be used to accelerate analysis and discovery in
a variety of application areas, ranging from surveillance to remote
sensing to ecosystem monitoring.  Many investigations employ cameras
to observe phenomena of interest.  By focusing attention on the most
novel images, our approach can help investigators quickly zero in on
the observations most likely to lead to new discoveries.
%- could be used to aid exploration and discovery in large scientific
%data sets as they are collected.

% If the authors' names are included in the sentence, place only
%the year in parentheses, for example when referencing Arthur Samuel's
%pioneering work \yrcite{Samuel59}. Otherwise place the entire
%reference in parentheses with the authors and year separated by a
%comma \cite{Samuel59}. List multiple references separated by
%semicolons \cite{kearns89,Samuel59,mitchell80}. Use the `et~al.'
%construct only for citations with three or more authors or after
%listing all authors to a publication in an earlier reference
%\cite{MachineLearningI}. 

%\subsection{Software and Data}
%We strongly encourage the publication of software and data with the
%camera-ready version of the paper whenever appropriate. This can be
%done by including a URL in the camera-ready copy. However, do not
%include URLs that reveal your institution or identity in your
%submission for review. Instead, provide an anonymous URL or upload
%the material as ``Supplementary Material'' into the CMT reviewing
%system. Note that reviewers are not required to look a this material
%when writing their review.

% Acknowledgments should only appear in the accepted version.
\section*{Acknowledgments}

We thank the Planetary Data System Imaging Node for funding this
project.  Part of this research was carried out at the Jet Propulsion
Laboratory, California Institute of Technology, under a contract with
the National Aeronautics and Space Administration.  

% Typically, this will include thanks to reviewers
%who gave useful comments, to colleagues who contributed to the ideas,
%and to funding agencies and corporate sponsors that provided financial
%support.

% LocalWords:  DEMUD CNN ImageNet mislabeled adversarial al CMT autoencoder

\bibliography{discovery}
\bibliographystyle{icml2018}

\end{document}